\documentclass{ifacconf}

\usepackage{amsmath,amssymb}
\usepackage{amsfonts}
\usepackage{graphicx}      
\usepackage{natbib}        
\usepackage{color}
\usepackage{xurl}
\def\MS#1{{\color{black}#1}}
\def\HJ#1{{\color{black}#1}}
\def\HJfinal#1{{\color{black}#1}}

\begin{document}
\begin{frontmatter}

\title{Model Predictive Planner for UAV Navigation in Non-Convex Air Corridors\thanksref{footnoteinfo}} 

\thanks[footnoteinfo]{This manuscript version is made available under the CC-BY-NC-ND 4.0 license. This work was supported in part by the Brazilian agencies CAPES through the Academic Excellence Program (PROEX), CNPq under the grants 317058/2023-1 and 422143/2023-5, and in part by Petrobras/ANP under Grants 2023/00494-5 and 2023/00643-0. Marcelo A. Santos acknowledges support from the Lombardy Region under the PR FESR 2021–2027 "Collabora \& Innova" call (Decree no. 11969, 2 August 2024), Project "HARMONY" (CUP: E59I25000850007).}

\author[First]{Henrique Silva, Jr.} 
\author[Second]{Marcelo A. Santos} 
\author[First,Third]{Guilherme V. Raffo}

\address[First]{Graduate Program in Electrical Engineering, Universidade Federal de Minas Gerais, Belo Horizonte, MG, Brazil}
\address[Second]{Department of Management, Information and Production Engineering, University of Bergamo, Dalmine, BG, Italy}
\address[Third]{Department of Electronic Engineering, Universidade Federal de Minas Gerais, Belo Horizonte, MG, Brazil}

\begin{abstract}               
\HJfinal{This work presents a motion planning framework for UAV navigation in non-convex urban air corridors. The planner is based on a mixed-integer tracking model predictive control formulation that enforces corridor feasibility and dynamic consistency within a single optimization problem. To guarantee convergence to the target and mitigate the occurrence of local minima induced by non-convex geometry, a shortest-path-based offset cost with feasibility constraints is embedded directly into the planning problem. Numerical simulations show that the proposed formulation generates dynamically valid trajectories that satisfy the corridor constraints and converge to the target without relying on external global planning stages.}

\end{abstract}


\end{frontmatter}

\section{Introduction}

\HJfinal{Driven by significant advances in autonomy and embedded computing, Unmanned Aerial Vehicles (UAVs) have become essential tools for civil applications, ranging from infrastructure inspection to logistics. This widespread adoption is closely tied to Urban Air Mobility (UAM) efforts to organize and manage autonomous flights in shared low-altitude airspace. To ensure operational safety and standardization, aviation authorities have issued regulatory frameworks, including the FAA/NASA ConOps \citep{FAA_UAM_ConOps_v2_2023}, EASA’s U-Space \citep{EASA_EAR_U-space_2024}, and ICAO’s UTM \citep{ICAO_UTM_Framework_2023}.}

\HJfinal{The constrained structure of UAM airspace makes motion planning a central requirement. A typical hierarchical approach \citep{uav_survey} employs a high-level planner to compute a geometrically feasible path via search or sampling techniques \citep{karur2021survey}. Subsequently, a dynamic control layer computes trajectories and inputs to track this path while enforcing system and safety requirements. Model Predictive Control (MPC) is a common choice for this layer \citep{rawlings2009model}, being extensively explored in navigation tasks \citep{Brito2019MPCC, drones6050126}. 
However, this framework relies on a strict time-scale separation \citep{layered}, limiting real-time adaptation in dynamic environments.}


\MS{Within this layered paradigm, an emerging alternative is to reinterpret predictive control as a motion planning mechanism rather than solely as a tracking controller. In this view, the planner directly optimizes dynamically feasible trajectories over a receding horizon, taking into account both the system dynamics and the admissible airspace constraints, while a separate low-level controller is retained for fast stabilization and disturbance rejection. This perspective preserves the hierarchical structure, but replaces purely geometric planning with a predictive, constraint-aware planning layer, in which dynamically feasible reference trajectories are generated in real-time.}

\HJfinal{Changing targets in UAM operations require continuous adaptation, motivating predictive formulations that unify motion generation and dynamic regulation \citep{santos2024}. This joint approach simultaneously addresses dynamic feasibility, constraint enforcement, and reference generation. However, standard MPC schemes may lose feasibility in this context, since their guarantees are typically local and hold only within a limited domain of attraction. This limitation motivated the tracking MPC framework \citep{limon2018}, which optimizes both the predicted trajectory and an artificial steady-state. By penalizing the deviation from the true target via an offset cost, tracking MPC accommodates target variations, preserves recursive feasibility, and enlarges the domain of attraction. While suitable for constrained navigation, its stability guarantees rely on convexity arguments \citep{tracking_tutorial}, a condition that rarely holds in UAM airspaces featuring protected regions and structured corridors.}


\HJfinal{Several strategies address non-convexity in MPC-based frameworks. Topological transformations \citep{homeomorfismo} map specific non-convex sets into convex ones, though their applicability is limited by the underlying set structure. Other approaches employ multi-trajectory mechanisms to avoid local minima, such as dual-trajectory schemes \citep{exploration} or LIDAR-based branching \citep{nascimento2023nmpc}. Alternatively, redefining the cost function can provide convergence guarantees in non-convex spaces. For instance, \citet{soloperto} introduced an offset cost based on the shortest feasible curve connecting an artificial reference to the target, an idea refined by \citet{milagroso} to improve tractability in cluttered environments.}

\HJfinal{This work develops a model predictive planning strategy for UAV navigation in constrained air corridors, modeled as a sequence of convex sets forming a non-convex region. Building on \citet{milagroso}, we embed a shortest-path proxy into a mixed-integer tracking MPC to ensure convergence in non-convex domains while preserving dynamic feasibility. Corridor boundaries are enforced via integer constraints, and a geometry-exploiting warm-start strategy improves computational performance. The main contributions are threefold: i) a predictive planning framework handling target convergence and non-convex constraints within a single optimization problem; ii) a shortest-path-inspired cost adapted for air corridors; and iii) a geometry-based warm start for real-time operation.}



\section{Problem Definition}
\begin{figure}[htbp!]
    \centering
    \includegraphics[width=1\linewidth]{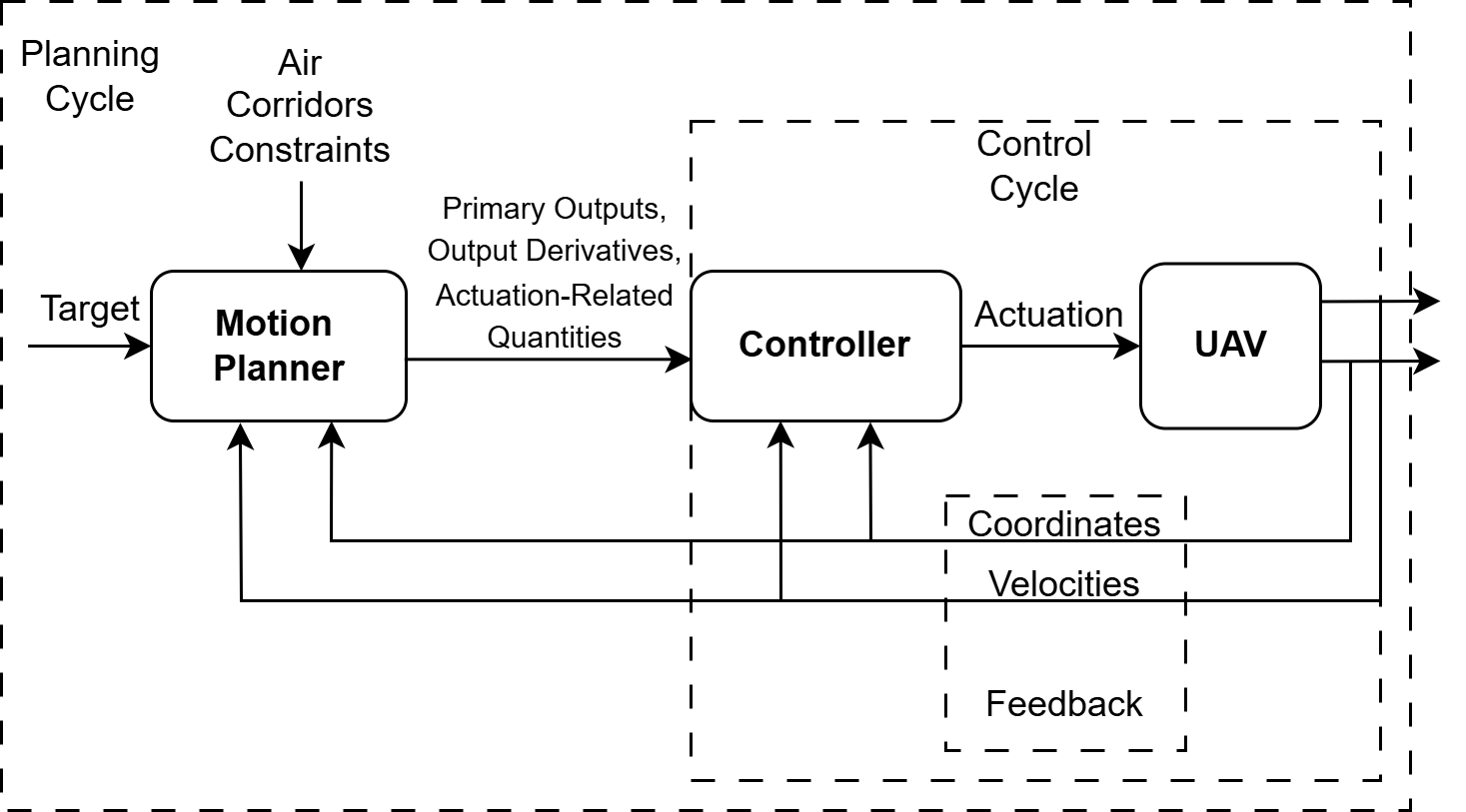}
    \caption{Planning and control scheme adopted for UAV navigation in structured UAM environments.}
    \label{fig:diagram}
\end{figure}

\HJfinal{This work addresses UAV navigation within structured UAM air corridors characterized by non-convex geometry. The planner must generate trajectories that remain within the admissible airspace while satisfying vehicle dynamics, ensuring the references supplied to the lower-level controller are dynamically feasible. As illustrated in Fig.~\ref{fig:diagram}, the motion planner generates these reference signals, which may include primary outputs, derivatives, or actuation quantities, tailored to the underlying control architecture.}


To avoid relying on purely geometric paths, the planner optimizes over trajectories consistent with the vehicle dynamics. To this end, let a general class of UAV systems be described by the discrete-time nonlinear dynamics 
\[x(k+1) = f(x(k), u(k)),\]
where $x(k) \in \mathbb{R}^n$ and $u(k) \in \mathbb{R}^m$ are the state and control input vectors, respectively. The prediction model $f(\cdot)$ and the composition of these vectors depend on the planning and control architecture requirements. In addition, the state and input must satisfy the operational constraints 
\[x(k), u(k)) \in \mathbb{X} \times \mathbb{U}, ~ \forall ~ k \ge 0,\]
\HJfinal{where the compact convex sets $\mathbb{X} \subset \mathbb{R}^n$ and $\mathbb{U} \subset \mathbb{R}^m$ reflect} physical limitations such as velocity bounds and actuator limits. Moreover, the output map is designed to describe the vehicle position, as 
\[y(k) = h\big(x(k)\big) \in \mathbb{Y}, ~ \forall ~ k \ge 0,\]
\HJfinal{where $\mathbb{Y} \subset \mathbb{R}^3$} denotes the navigable workspace.

The admissible airspace $\mathbb{Z} \subset \mathbb{Y}$ is modeled as a non-convex union of \HJ{$n_z$} convex corridor segments, \HJfinal{$\mathbb{Z} = \bigcup_{i=1}^{n_z} \mathbb{Z}_i ,$} where the corridor segments $\mathbb{Z}_i$ are assumed to intersect each other only at their borders and each region is represented using a zonotopic description. This choice is motivated by the fact that zonotopes provide a compact parametrization of convex sets and support efficient containment checks within optimization problems. Then, each $\mathbb{Z}_i$ is given by
\begin{equation} \label{eq:zonotope_definition}
    \mathbb{Z}_i = \{\, c_i + G_i \xi \;\mid\; \xi \in [-1,1]^{p_i} \,\},
\end{equation}
with center $c_i \in \mathbb{R}^3$ and matrix $G_i \in \mathbb{R}^{3\times p_i}$ \HJ{containing $p_i$ generator vectors}. 

To support model predictive planning in this non-convex domain, consider an artificial steady-state pair $(x_s,u_s)$ and its corresponding output $y_s$ satisfying 
\[x_s = f(x_s,u_s) \text{ and } y_s = h(x_s).\]
Then, the admissible set of steady states and inputs is defined as 
\begin{equation}
    \mathbb{S}
    = \{\, (x_s,u_s) \in \lambda(\mathbb{X} \times \mathbb{U})
        \;|\;
        x_s = f(x_s,u_s) \,\},
\end{equation}
where $\lambda \in (0,1)$ is a scalar arbitrarily close to one, introduced to avoid equilibrium points active at the feasibility constraints \citep{limon2018}.  It is worthwhile mentioning that the artificial output $y_s$ acts as an intermediate reference optimized online, ensuring that the planner selects a target that is dynamically consistent and remains inside the admissible airspace.

Based on these elements, the motion planning problem is formulated as a tracking MPC scheme:
\begin{subequations} \label{eq:otimizacao_compacta}
\begin{align}
\min_{\textbf{u}, \textbf{x}, u_s, x_s} \quad & \sum_{j=0}^{N-1} \ell(x(j)-x_s, u(j)-u_s) + V_{o}(y_s, y_t)\nonumber\\
\text{s.t.} \qquad & x(0)=x, \label{eq:cond_inicial} \\
& x(j+1) = f(x(j), u(j)), \quad j \in \mathbb{N}_{0:N-1}, \label{eq:dinamica} \\
& (x(j), u(j)) \in \mathbb{X} \times \mathbb{U}, \quad j \in \mathbb{N}_{0:N}, \label{eq:limites_operacionais} \\
& y(j) \in \mathbb{Z}, \quad j \in \mathbb{N}_{0:N},  \label{eq:rest_zon_1} \\
& (x_s, u_s) \in \mathbb{S},  \label{eq:restricoes_steady} \\
& y_s \in \mathbb{Z},  \label{eq:rest_ys_z}\\
& x(N) = x_s. \label{eq:restricao_terminal}
\end{align}
\end{subequations}
\HJfinal{Here, $\ell(\cdot)$ is a positive definite stage cost guiding the predicted trajectory towards the artificial state, while the offset cost $V_{o}(\cdot)$ penalizes the deviation between the artificial output and the real target $y_t$. Constraints \eqref{eq:cond_inicial} and \eqref{eq:dinamica} enforce the initial condition and system dynamics. Moreover, \eqref{eq:limites_operacionais} and \eqref{eq:rest_zon_1} impose operational bounds and limit the predicted output to the admissible airspace, respectively. The artificial variables and their spatial admissibility are established by \eqref{eq:restricoes_steady} and \eqref{eq:rest_ys_z}, while \eqref{eq:restricao_terminal} guarantees terminal convergence to the artificial reference. Finally, $N$ denotes the prediction horizon and $\mathbb{N}_{a:b}$ the set of integers in $[a,b]$.}

\section{Non-Convex Corridor Constraints and Global Convergence}
This section presents the modeling of the corridor constraints together with the formulation proposed to achieve global convergence based on a shortest-path offset cost.

\subsection{Zonotopic Corridor Constraints} \label{subsec:zonotopic_constraints}
Since the admissible corridor is modeled as a non-convex union of convex segments, the containment requirement in \eqref{eq:rest_zon_1} and \eqref{eq:rest_ys_z} must be enforced for every predicted output along the horizon \MS{and the artificial output}. For each corridor segment $\mathbb{Z}_i$, defined in \eqref{eq:zonotope_definition}, verifying that a point $y$ belongs to this set can be performed through different approaches, including quadratic programming \citep{combastel2013quadratic} and H-representations \citep{althoff2010efficient}. In this work, we consider the containment verification $y \in \mathbb{Z}_i$ simply as a feasibility problem, expressed as 
\[\exists \xi_i \in [-1, 1]^{p_i} \quad \text{s.t.} \quad y - c_i - G_i \xi_i = 0.\]

Since each predicted output must belong to exactly one corridor segment at every prediction step, we introduce binary decision variables to encode this assignment over the horizon. Therefore, we define 
\[\delta_i(j) \in \{0,1\}, ~ \forall i \in \mathbb{N}_{1:n_z} \text{ and } \forall j \in \mathbb{N}_{0:N}.\]

This conditional activation can be implemented in an optimization problem using a Big-M formulation \citep{bemporad1999control}, which links the output $y$ to the active corridor segment. Thus, the predicted output feasibility constraints of (9) are reformulated as
\begin{equation} \label{eq: bigM}
    \|y(j)-c_i-G_i\xi_i(j)\|_\infty\leq M(1-\delta_i(j)),
\end{equation}
for all $i \in \mathbb{N}_{1:n_z}$ and $j \in \mathbb{N}_{0:N}$, where $M$ is a sufficiently large positive constant and $\|\cdot\|_\infty$ denotes the infinity norm. Finally, the exclusive assignment of each predicted output to a single segment is enforced through 
\[\sum_{i=1}^{n_z} \delta_i(j) = 1, ~ \forall j \in \mathbb{N}_{0:N}.\]

It is worth noting that no zonotopic reformulation is required for the artificial output, since the terminal condition $x(N) = x_s$ implicitly ensures that $y_s \in \mathbb{Z}$. Indeed, once $y(N) \in \mathbb{Z}$ is enforced, the terminal equality implies $y_s = h(x_s) = h(x(N))$.

\subsection{Shortest-Path Offset Cost for Global Convergence}
\HJfinal{Standard tracking MPC schemes typically use an Euclidean distance-based offset cost, which can lead to local minima and prevent global convergence in non-convex domains. To mitigate this, \cite{milagroso} replaces the Euclidean measure with a shortest-feasible-path construction that accounts for the admissible space topology. The formulation proposed in this work adapts this concept to the corridor structure defined in Section~2, ensuring target convergence even in non-convex airspace.}


To represent the path connecting the artificial output $y_s$ to the target $y_t$, consider a sequence of points $\boldsymbol{r}$ that forms $n_s$ straight segments. Since each segment must remain inside the admissible corridor, a discretized interpolation scheme is introduced. For that, let $N_p$ denote the number of interpolation steps and define 
\[\mathbb{A} = \{k/N_p | \; k \in \mathbb{N}_{0:{N_p}}\}.\]

For every segment $q \in \mathbb{N}_{0:n_s-1}$ and interpolation factor $\alpha \in \mathbb{A}$, the intermediate points are given by
\begin{equation}  \label{eq:interpol_points}
    r_p(q,\alpha) = (1-\alpha)\,r(q) + \alpha\,r(q+1),
\end{equation}
and must satisfy a zonotopic containment condition. 

As in Section~3.1, containment is enforced using auxiliary continuous variables $\gamma_i(q,\alpha)$, which encode the zonotopic generator coefficients, together with binary variables $\zeta_i(q,\alpha)$ that activate the corresponding corridor segment. Their relation is imposed through 
\begin{equation} \label{eq:zonotopo caminho_of_priori}
    \HJfinal{\|r_p -c_i-G_i\gamma_{i}\|_\infty \leq M(1-\zeta_{i}).}
\end{equation}

Moreover, the resulting offset cost is given by
\begin{equation}
    V_{o}(\boldsymbol{r}) = \kappa_s \sum_{q=0}^{n_s-1} \|\, r(q+1) - r(q) \,\|^2,
\end{equation}
where $\kappa_s$ weights the contribution of the path length. Then, the artificial path $\boldsymbol{r}$ is defined as the solution of the optimization problem
\begin{subequations} \label{eq:custo_offset_completo}
\begin{align} 
\min_{\boldsymbol{r}} ~ & V_{o}(\boldsymbol{r}) \nonumber\\
\text{s.t.} ~ &r(0) = y_s, \label{of_rest_1} \\
& r(n_s) = y_t, \label{of_rest_3} \\
& \text{\eqref{eq:interpol_points}}, \: q \in \mathbb{N}_{0:n_s-1}, \: \alpha \in \mathbb{A}, \label{eq:yp}\\
&\text{\eqref{eq:zonotopo caminho_of_priori}}, \: q \in \mathbb{N}_{0:n_s-1}, \: \alpha \in \mathbb{A}, \label{zonotopo caminho_of}\\ 
& \sum_{i=1}^{n_z} \zeta_i(q, \alpha) = 1, \; q \in \mathbb{N}_{0:n_s-1}, \: \alpha \in \mathbb{A}.\label{eq:exclusive_of}
\end{align}
\end{subequations}

\begin{figure}[bp!]
    \centering
    \begin{minipage}[b]{0.48\columnwidth}
        \centering
        \includegraphics[width=\linewidth]{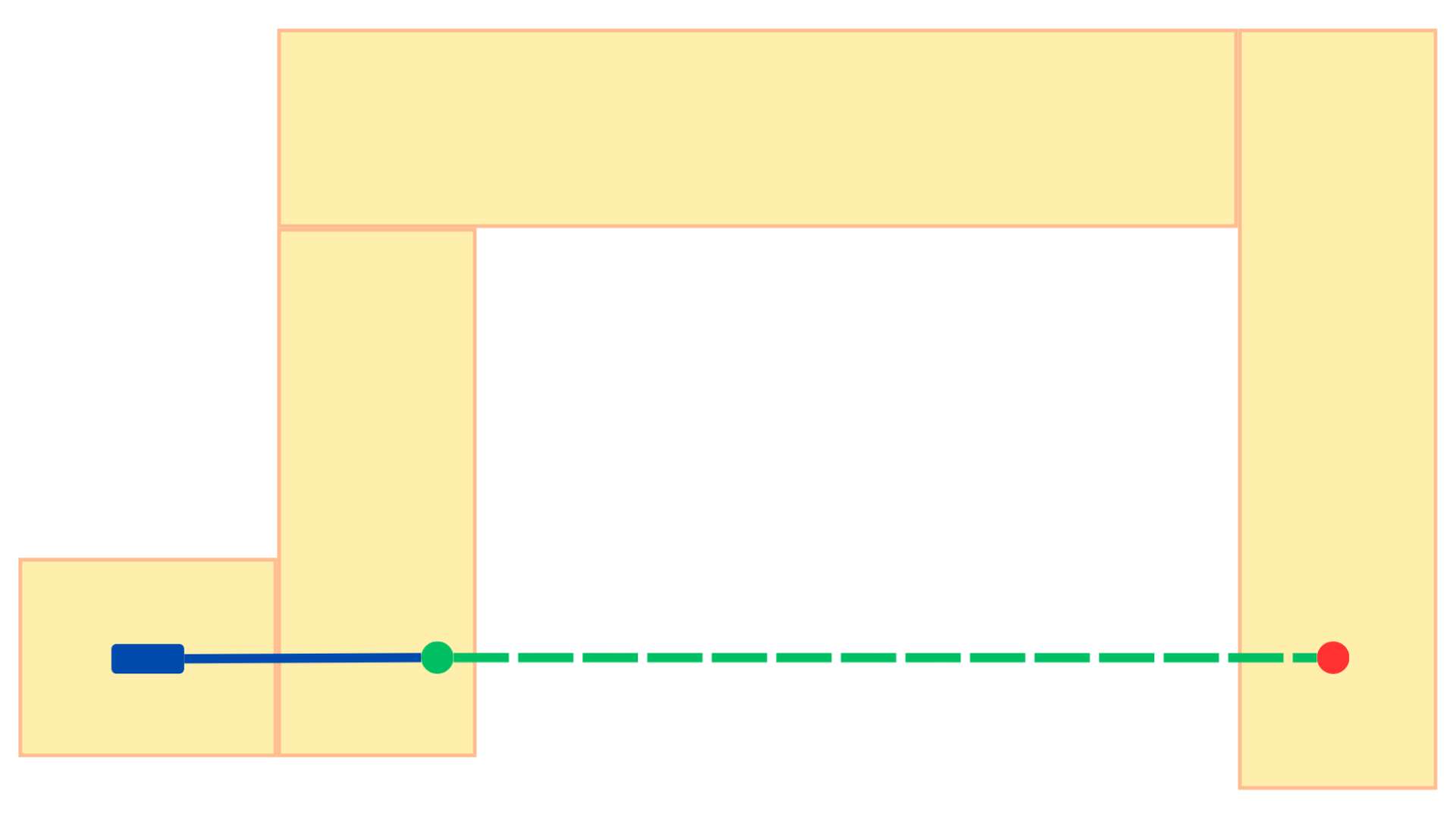}
        \centerline{\footnotesize (a) Standard approach} \label{fig:offset1}
    \end{minipage}
    \hspace{0.01\columnwidth} 
    \begin{minipage}[b]{0.48\columnwidth}
        \centering
        \includegraphics[width=\linewidth]{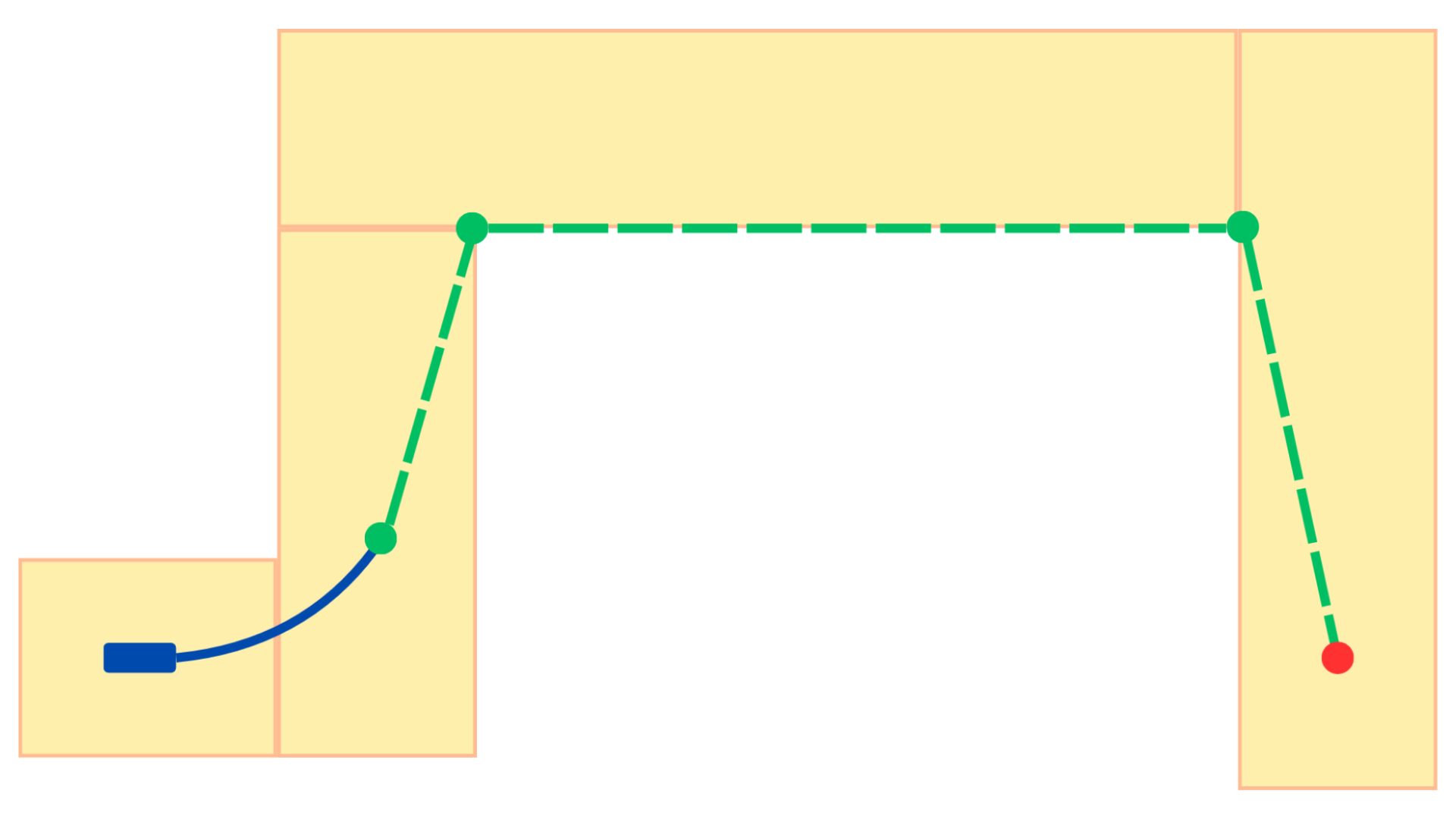}
        \centerline{\footnotesize (b) Proposed approach} \label{fig:offset2}
    \end{minipage}
    \vspace{-4mm}
    \caption{\footnotesize Illustration of the difference between the standard Tracking MPC formulation and the approach proposed in this work. The airspace delimited by zonotopes is represented in yellow, and the target in red. The UAV (blue box) optimizes the predicted state trajectory (blue segment) towards the artificial reference (green circle), which in turn optimizes the offset cost (green dashed line) towards the target. In (a), the standard offset cost causes the system to get stuck in a local minimum, whereas in (b), convergence to the target is achieved.}
    \label{fig:offset}
\end{figure}

\HJfinal{This optimization problem computes the shortest feasible $n_s$-segment path anchored between $y_s$ and $y_t$ via \eqref{of_rest_1} and \eqref{of_rest_3}. The discretization \eqref{eq:yp} enforces the containment constraints \eqref{zonotopo caminho_of} and \eqref{eq:exclusive_of} only on a finite set of interpolated points, avoiding the prohibitive cost of checking continuous segments. Thus, the formulation captures the non-convex connectivity while remaining tractable. By minimizing this piecewise-linear path, the offset cost induces global convergence, effectively guiding the artificial reference toward the target, as illustrated in Fig.~\ref{fig:offset}.}



\section{Model Predictive Planning Formulation}
Considering the tracking MPC formulation in (9) together with the mixed-integer corridor constraints of Section 3.1 and the shortest-path offset cost of Section 3.2, the planning problem can be posed as a single mixed-integer optimization problem. This formulation simultaneously obtains the predicted trajectory, the artificial variables, and the geometric connection to the target while enforcing the admissible airspace constraints, and is given by
\begin{subequations} \label{eq:otimizacao_completa}
\begin{align}
\min_{\substack{
\mathbf{u},\ \mathbf{x},\ \mathbf{r},\ \boldsymbol{\delta},\ \boldsymbol{\xi},\\
\boldsymbol{\gamma},\ \boldsymbol{\zeta},\ u_s,\ x_s
}} & \;\; \sum_{j=0}^{N-1} \ell(x(j)-x_s, u(j)-u_s) + V_o(\boldsymbol{r}) \nonumber\\
\text{s.t.} ~~\quad &x(0)=x,\\
&x(j+1) = f(x(j), u(j)), \quad j \in \mathbb{N}_{0:N-1},\\
&(x(j), u(j)) \in \mathbb{X} \times \mathbb{U}, \quad j \in \mathbb{N}_{0:N-1}, \label{modelo}\\
&\text{\eqref{eq: bigM}}, \quad i \in \mathbb{N}_{1:n_z}, \quad j \in \mathbb{N}_{0:N},\label{zonotopo trajetoria} \\ 
& (x_s, u_s) \in \mathbb{S}, \label{ponto de equilibrio} \\
& x(N) = x_s, \label{restricao_terminal} \\
& r(0) = y_s, \label{restricao inicial offset} \\
&\text{\eqref{eq:interpol_points}}, \: q \in \mathbb{N}_{0:n_s-1}, \: \alpha \in \mathbb{A}, \label{eq:yp_1} \\
&\text{\eqref{eq:zonotopo caminho_of_priori}}, \: q \in \mathbb{N}_{0:n_s-1}, \: \alpha \in \mathbb{A}, \\ 
& r(n_s) = y_t, \label{final caminho} \\
&\sum_{i=1}^{n_z} \delta_i(j) = 1, \quad j \in \mathbb{N}_{0:N},\label{soma delta} \\
&\sum_{i=1}^{n_z} \zeta_i(q,\alpha) = 1, \quad q \in \mathbb{N}_{0:n_s-1}, \: \alpha \in \mathbb{A}, \label{soma zeta}
\end{align}
\end{subequations}
where $\ell(\cdot) = \|x(j) - x_s\|^{2}_{Q} + \|u(j)-u_s\|^{2}_R$ with the matrices $Q \geq 0$ and $R > 0$ representing the weighting factors on the states and the inputs, respectively. The decision variables include the predicted state and input sequences $\mathbf{x}$ and $\mathbf{u}$, the artificial reference $(x_s,u_s)$, the artificial path nodes $\mathbf{r}$, and the auxiliary continuous variables $\boldsymbol{\xi}$ and $\boldsymbol{\gamma}$ together with the binary variables $\boldsymbol{\delta}$ and $\boldsymbol{\zeta}$, which together encode the mixed-integer zonotopic constraints. Finally, the initial condition $x$ is a parameter of the problem.

\HJfinal{In the optimization problem \eqref{eq:otimizacao_completa}, the cost functional jointly optimizes the predicted trajectory toward the artificial reference while simultaneously steering this reference through the non-convex space toward the target. Constraints (9a)-(9c) establish the initial condition, system dynamics, and operational bounds, whereas (9d) restricts the predicted outputs to the admissible corridor. The artificial equilibrium is defined by (9e) and enforced as a terminal equality constraint in (9f). The artificial path is anchored between $y_s$ and $y_t$ via (9g) and (9j), with its feasibility verified through the interpolation rule (9h) and zonotopic containment constraints (9i). Finally, binary summation constraints (9k) and (9l) ensure the exclusive selection of corridor segments.}


In a receding-horizon setting, the optimization problem in~(19) is solved at each step using the current state as the initial condition. This iterative structure updates the predicted motion as the UAV progresses, providing a direct mechanism to accommodate changes in the target or in the admissible corridor while remaining consistent with the operational constraints. The formulation thereby supports real-time adaptation within the constrained airspace.

\subsection{Initialization Strategy}
\begin{figure}[htbp!]
    \centering
    \begin{minipage}[b]{0.32\columnwidth}
        \centering
        \includegraphics[width=\linewidth]{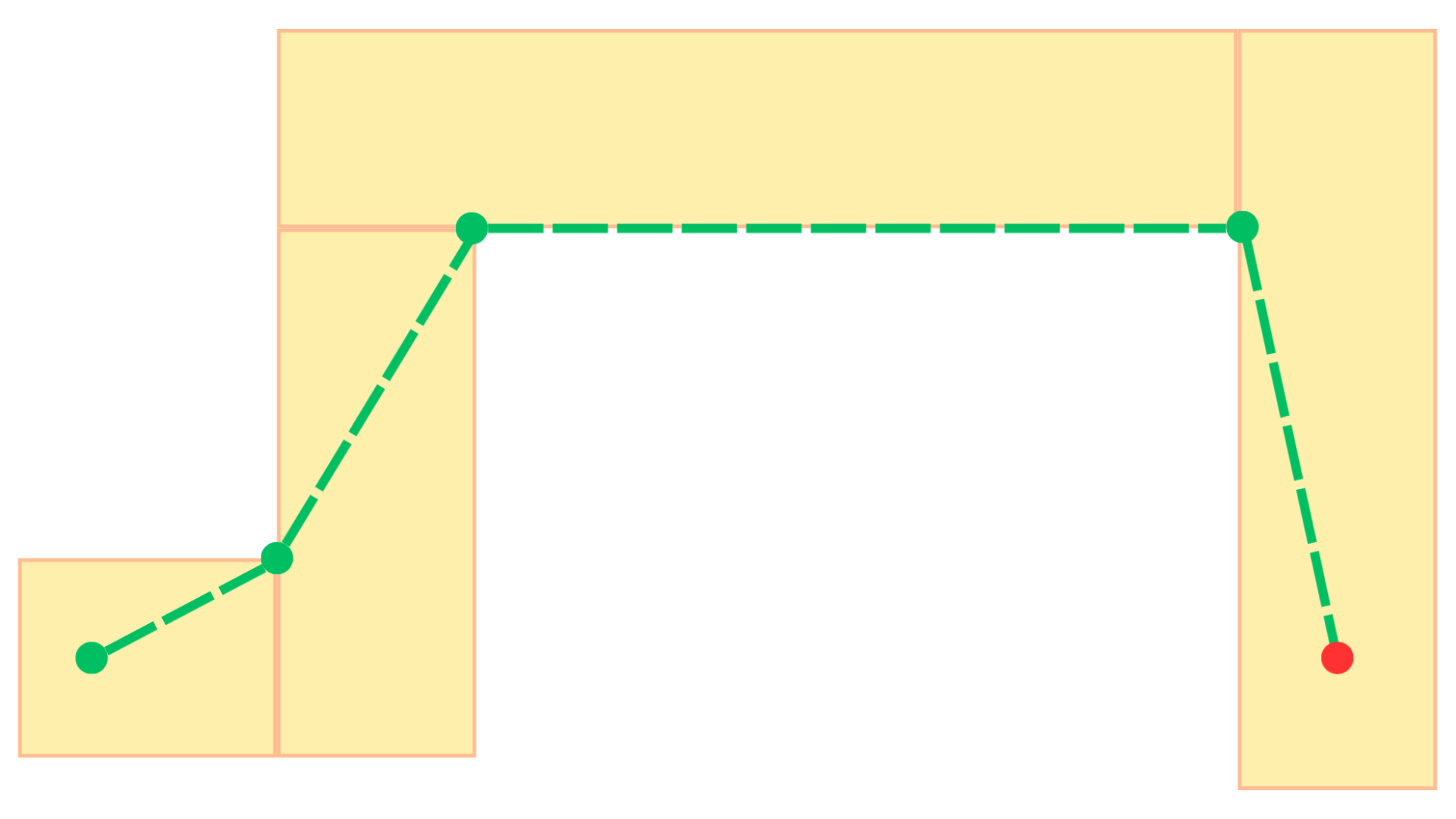}
        \centerline{\footnotesize (a) $n_s=n_z$} \label{fig:warm1}
    \end{minipage}
    \hfill
    \begin{minipage}[b]{0.32\columnwidth}
        \centering
        \includegraphics[width=\linewidth]{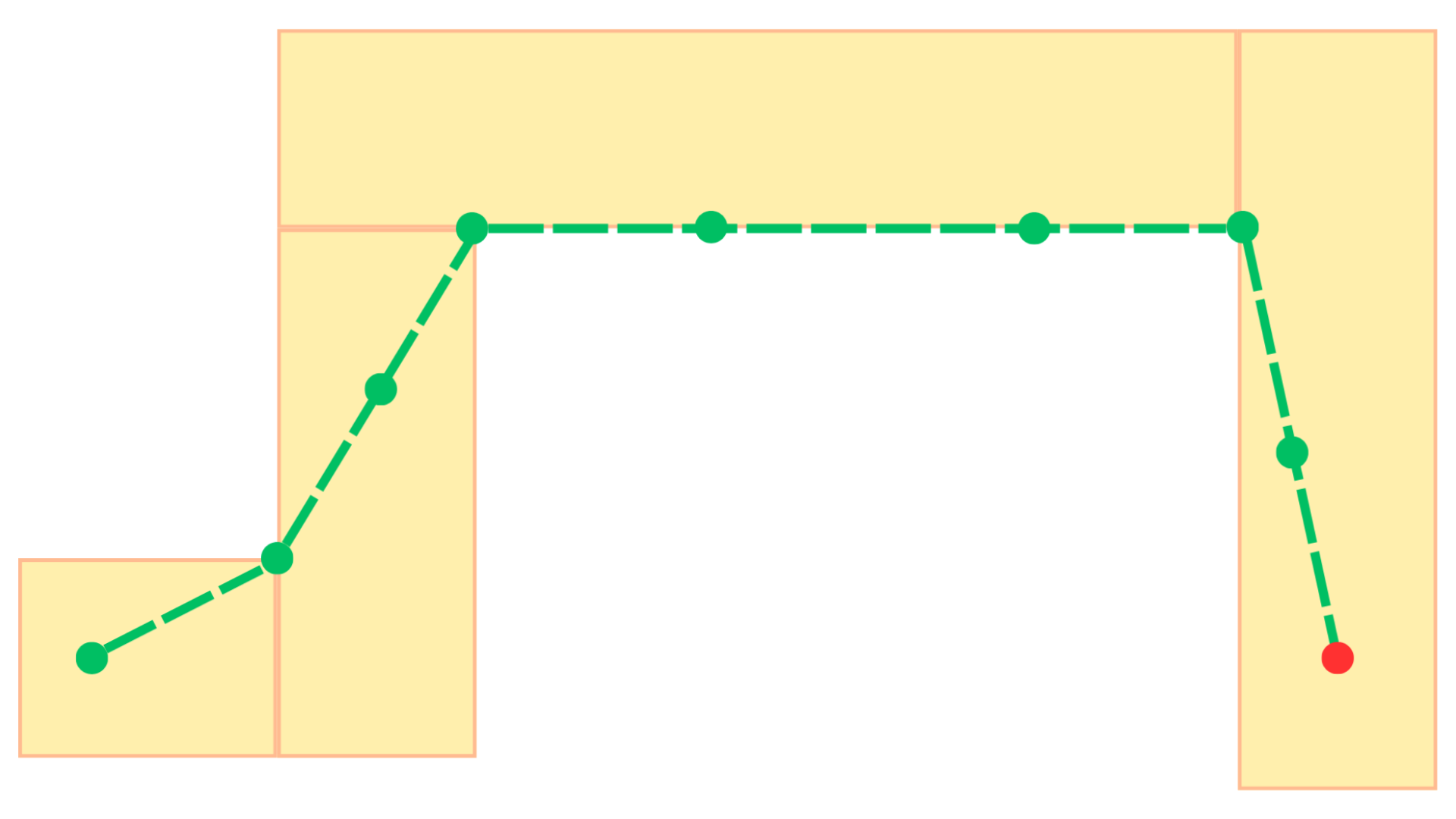}
        \centerline{\footnotesize (b) $n_s>n_z$} \label{fig:warm2}
    \end{minipage}
    \hfill 
    \begin{minipage}[b]{0.32\columnwidth}
        \centering
        \includegraphics[width=\linewidth]{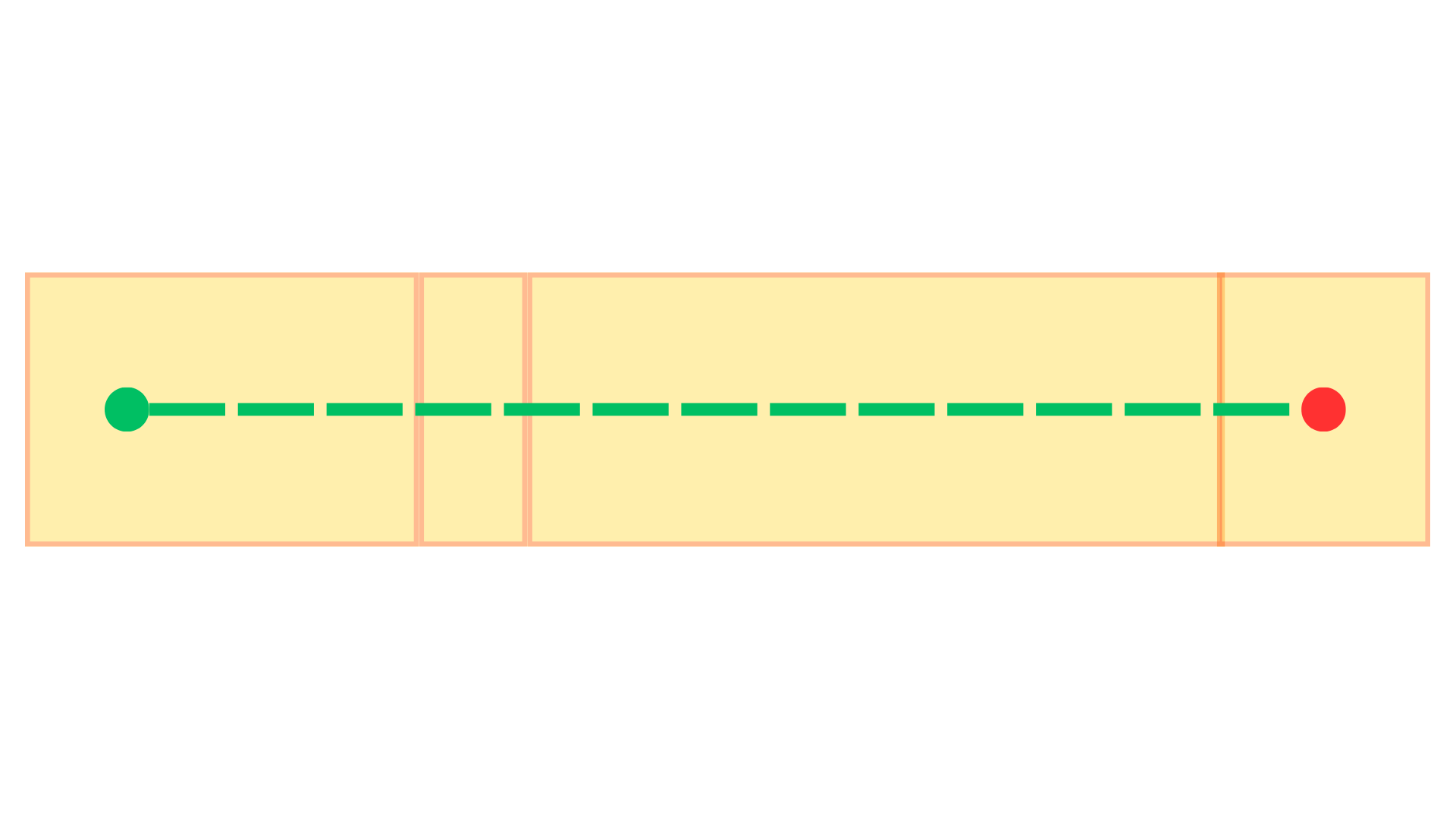}
        \centerline{\footnotesize (c) $n_s<n_z$} \label{fig:warm3}
    \end{minipage}
    \vspace{-4mm}
    \caption{\footnotesize Analysis of the number of path segments ($n_s$) relative to the number of zonotopes ($n_z$). In (a), $n_s = n_z$ is always sufficient to represent the shortest path. In (b), $n_s > n_z$ results in redundant segments. In (c), fewer segments ($n_s < n_z$) may be adequate for simpler topologies only.}
    \label{fig:warm_ns}
\end{figure}

\begin{figure}[htbp!]
    \centering
    \begin{minipage}[b]{0.43\columnwidth}
        \centering
        \includegraphics[width=\linewidth]{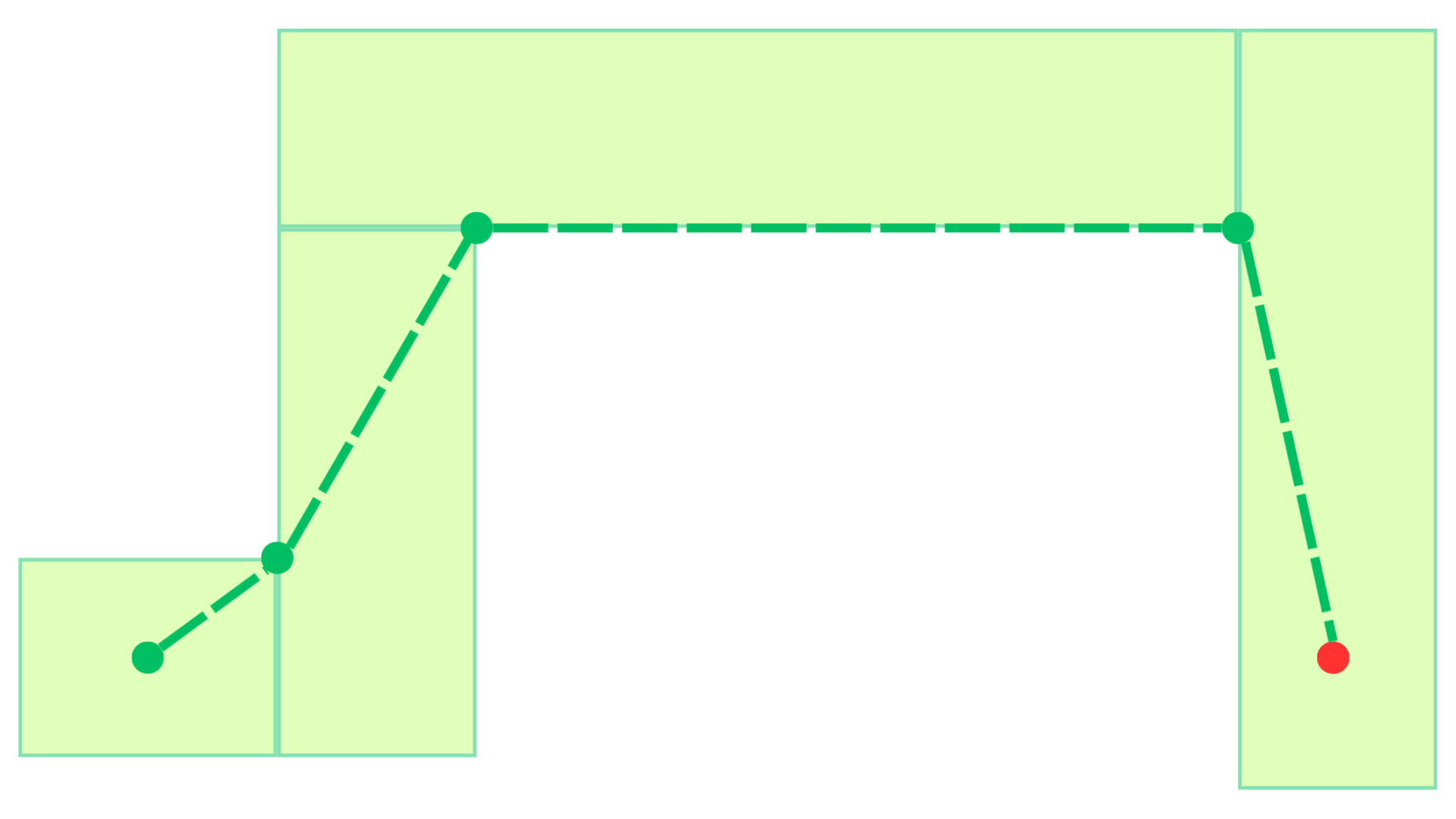}
        \centerline{\footnotesize (a)} 
    \end{minipage}
    \hspace{0.01\columnwidth}
    \begin{minipage}[b]{0.43\columnwidth}
        \centering
        \includegraphics[width=\linewidth]{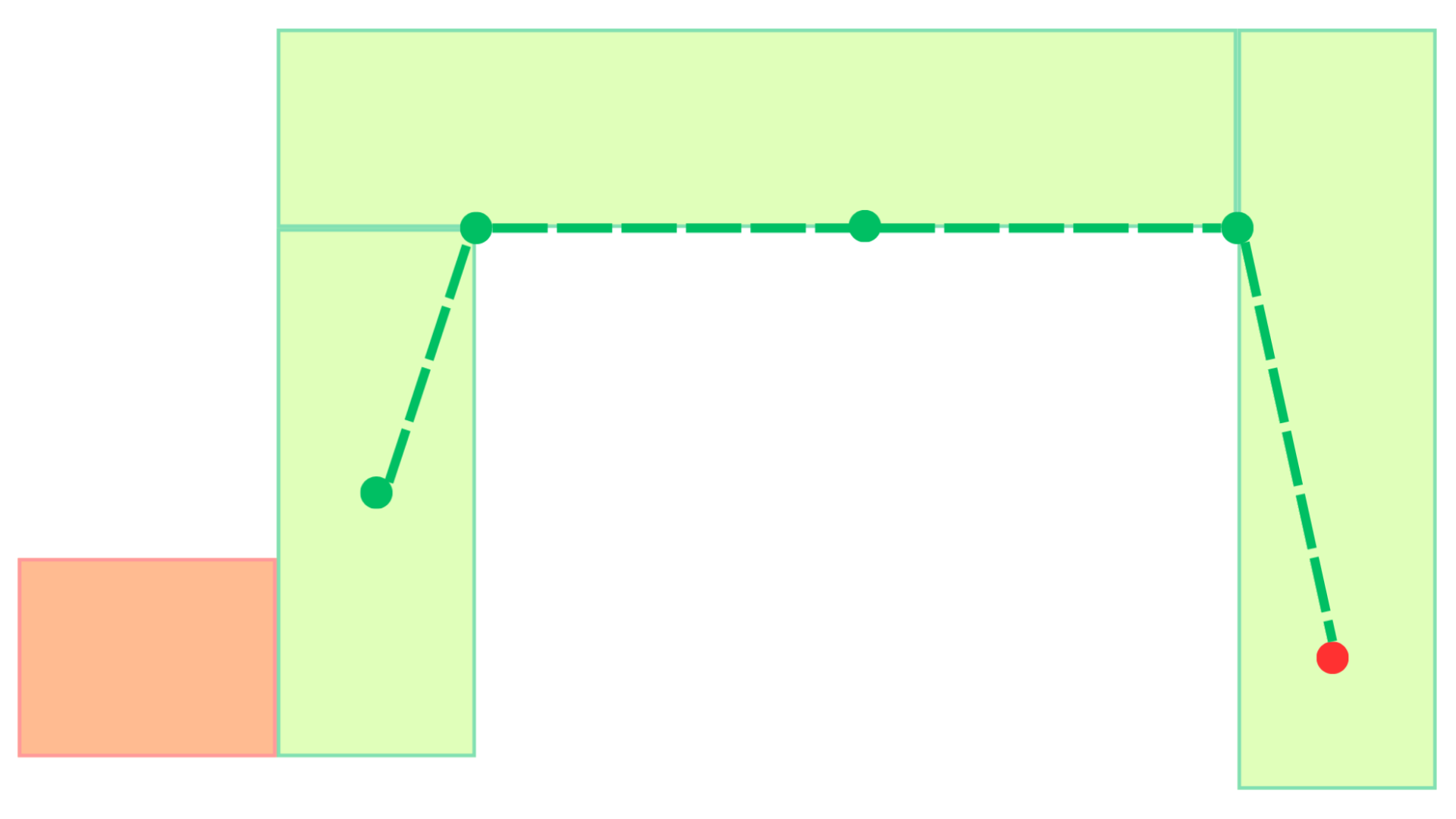}
        \centerline{\footnotesize (b)} 
    \end{minipage}
    \hspace{0.01\columnwidth}
    \begin{minipage}[b]{0.43\columnwidth}
        \centering
        \includegraphics[width=\linewidth]{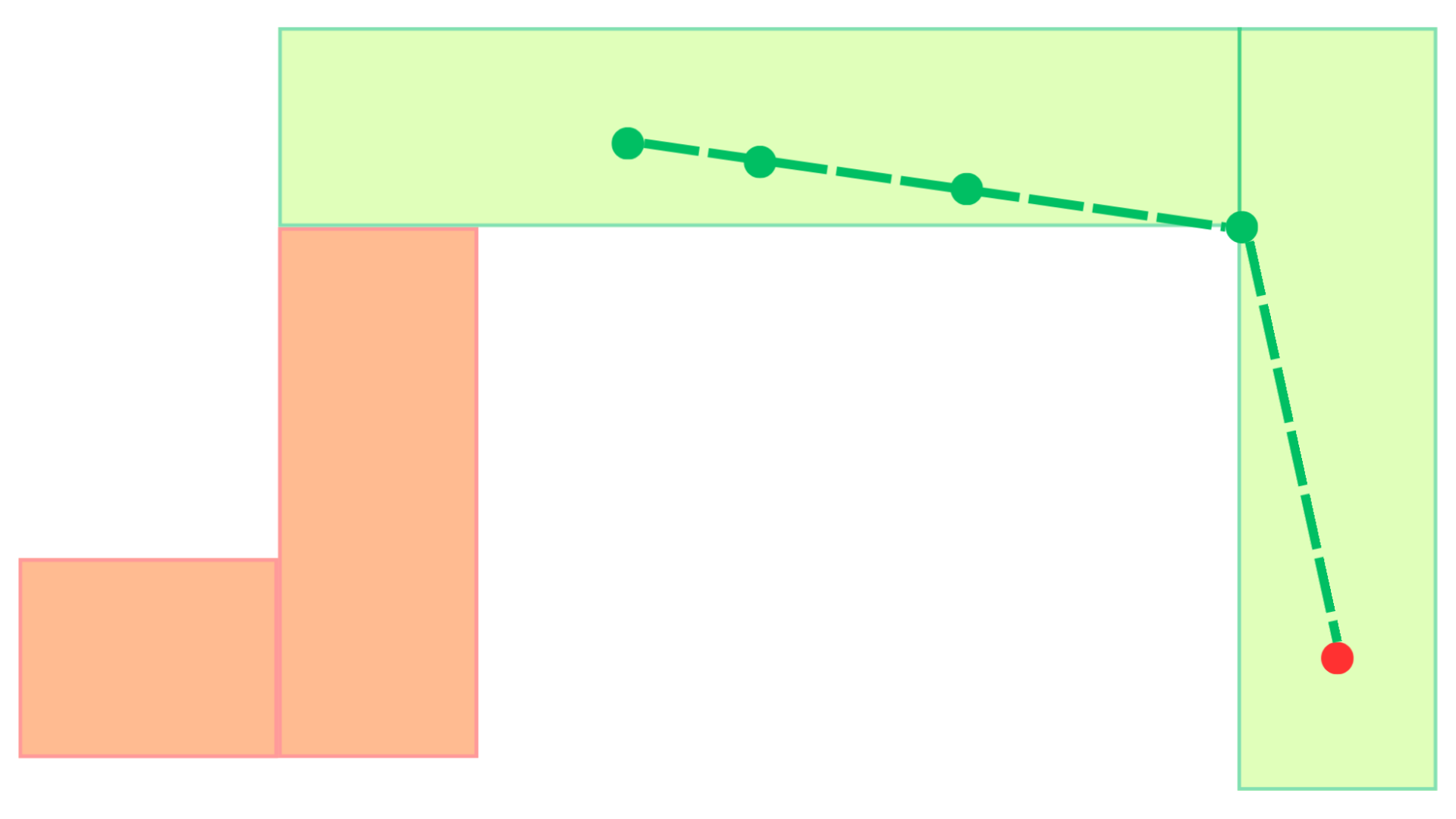}
        \centerline{\footnotesize (c)} 
    \end{minipage}
    \hspace{0.01\columnwidth}
    \begin{minipage}[b]{0.43\columnwidth}
        \centering
        \includegraphics[width=\linewidth]{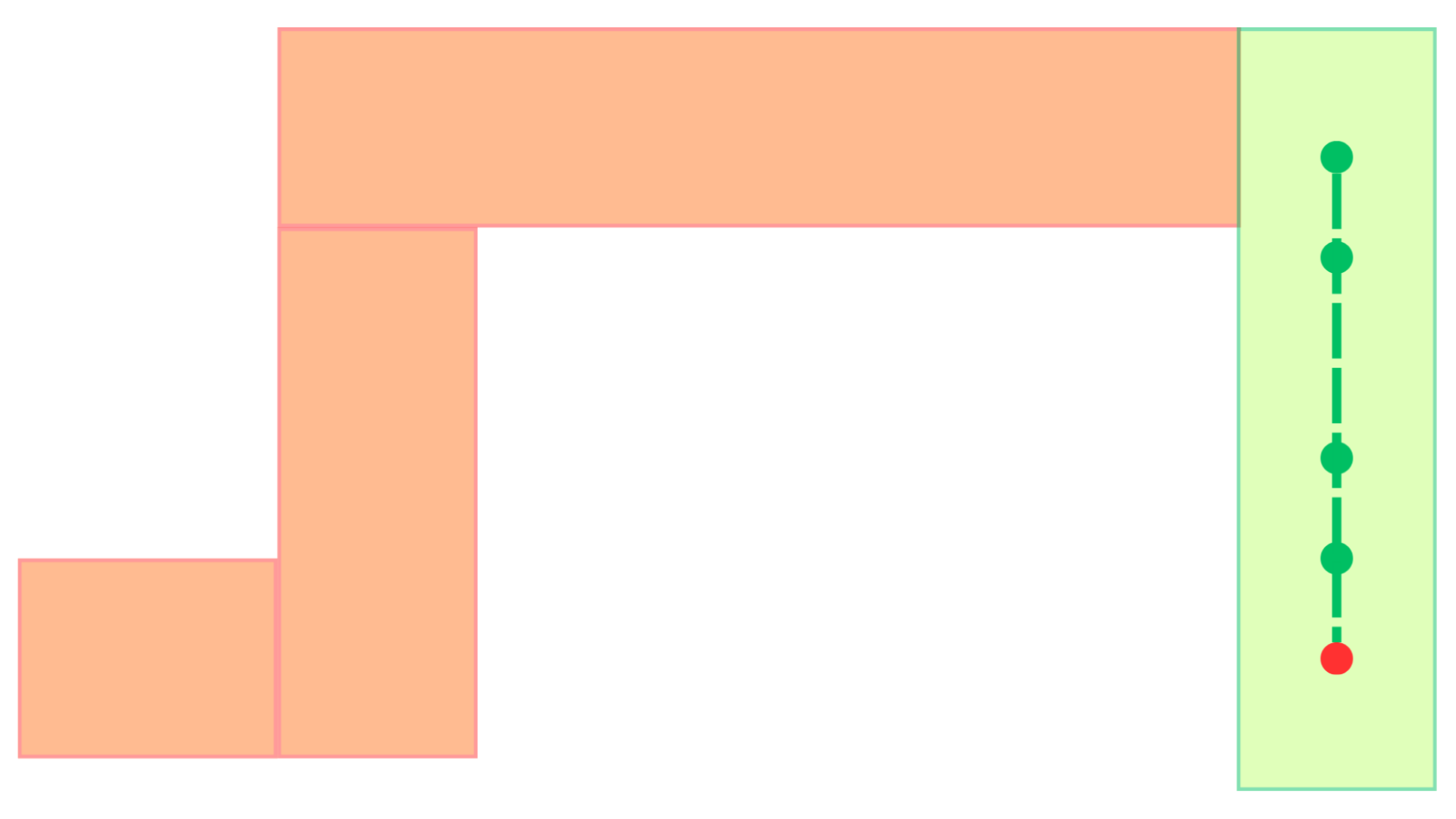}
        \centerline{\footnotesize (d)} 
    \end{minipage}
    \vspace{-2mm}
    \caption{\footnotesize Trajectory evolution from (a) to (d). The parameter $n_s$ is dimensioned for the most constrained scenario, which is the initial state (a). As the UAV progresses, the fixed $n_s$ becomes increasingly redundant but maintains feasibility without requiring dynamic adjustment.}
    \label{fig:warm_redund}
\end{figure}

\HJfinal{The offset mechanism embedded in \eqref{eq:otimizacao_completa} requires solving a mixed-integer problem whose convergence is highly sensitive to the initial iterate, making a suitable initialization essential for the first MPC iteration. A key design parameter in this setup is $n_s$, the number of linear segments used to represent the artificial path. The geometry of the corridor provides a natural criterion for selecting this value. Since each corridor region is modeled as a convex zonotope, a path that moves through a sequence of $n_z$ regions can be parameterized by $n_z$ linear segments, as illustrated in Fig.~\ref{fig:warm_ns}. Thus, $n_s = n_z$ is adopted, providing a representation consistent with the corridor structure. Although this choice may introduce redundancy as the UAV advances along the corridor, it ensures feasibility in the most restrictive portion of the trajectory, as illustrated in Fig.~\ref{fig:warm_redund}.}

To generate the initial guess for the first iteration, an auxiliary version of the optimization problem \eqref{eq:custo_offset_completo} is introduced, formulated as
\begin{subequations}
\begin{align}
\min_{\textbf{r}} \quad & V_o(\boldsymbol{r}) \nonumber \\
\text{s.t.} \quad & r(0) = y_0,  \\
& r(j) \in \{\mathbb{Z}_j, \: \mathbb{Z}_{j+1}\}, \quad j \in \mathbb{N}_{1:ns-1}, \label{eq:warm_rest_2}\\
& r(n_s) = y_t,
\end{align}
\end{subequations}
where $y_0$ corresponds to the UAV position at the start of the planning procedure.

\begin{figure}[htbp!]
    \centering
    \begin{minipage}[b]{0.43\columnwidth}
        \centering
        \includegraphics[width=\linewidth]{Figures/Sec7/warm1.png}
        \centerline{\footnotesize (a) With \eqref{eq:yp} and \eqref{zonotopo caminho_of}}
    \end{minipage}
    \hspace{0.02\columnwidth}
    \begin{minipage}[b]{0.43\columnwidth}
        \centering
        \includegraphics[width=\linewidth]{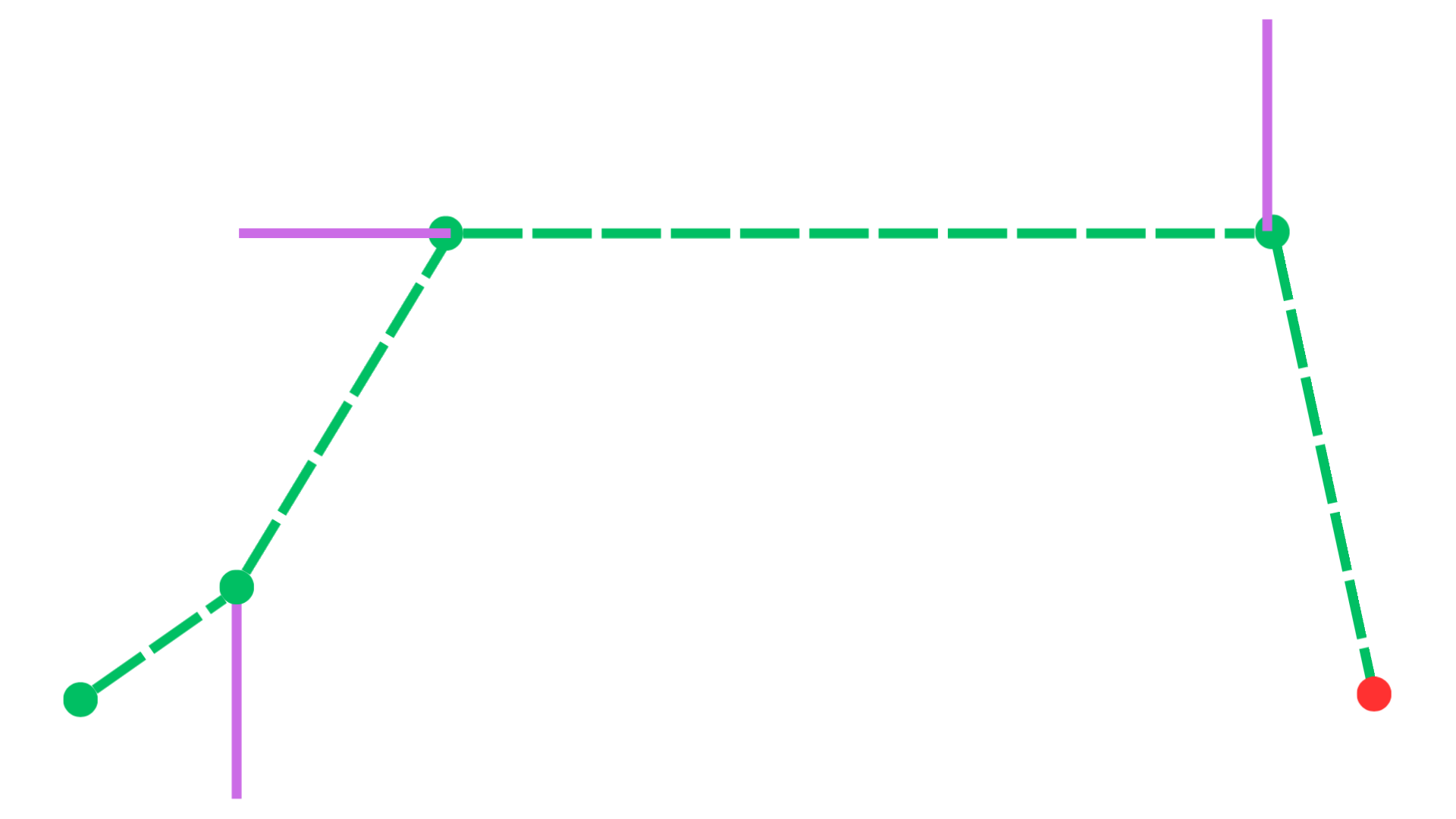}
        \centerline{\footnotesize (b) With \eqref{eq:warm_rest_2}} 
    \end{minipage}
    \caption{\footnotesize Analysis of the difference in the admissible space defined in the original problem (18), yellow sets, and in the auxiliary optimization problem (20), purple lines.}
    \label{fig:warm_rest}
\end{figure}

In this variation, the artificial path starts at the initial position, and the segment constraints \eqref{eq:yp} and \eqref{zonotopo caminho_of} are replaced by \eqref{eq:warm_rest_2}. The constraint \eqref{eq:warm_rest_2} significantly reduces the admissible search space by confining the path nodes to the intersections between consecutive zonotopes\HJ{, as shown in Fig. \ref{fig:warm_rest}}, yielding a more structured and easier initialization problem than the general formulation. Besides, following the zonotope definition, \eqref{eq:warm_rest_2} can be easily embedded in the optimization problem as a feasibility constraint. Furthermore, this initialization strategy removes the need for an external global planning stage or a precomputed roadmap, as required in \citet{milagroso}.

\section{Numerical Results}

\HJfinal{This section presents simulations to evaluate the planner's ability to satisfy non-convex constraints, manage corridor transitions, and converge to the target in a UAM corridor. All computations were carried out in Matlab 2025a using CasADi and the Bonmin mixed-integer solver.}

\HJfinal{The planning model is tailored to the low-level controller, which regulates and bounds position and linear velocity. Consequently, the planning state is defined as $x = [X,\, Y,\, Z,\, U,\, V,\, W]^\top$, allowing these constraints to be directly imposed and yielding admissible reference trajectories. The control inputs are the translational forces $u = [F_x,\, F_y,\, F_z]^\top$, consistent with a point-mass vehicle description. This ensures the planned motion respects low-level actuation limits, as force bounds directly translate into acceleration bounds.} 


Under this point-mass representation, and with sampling time $\tau_s$ and mass $m$, the discrete-time translational dynamics are
\begin{equation}
\begin{bmatrix}
X(k\!+\!1) \\
Y(k\!+\!1) \\
Z(k\!+\!1) \\
U(k\!+\!1) \\
V(k\!+\!1) \\
W(k\!+\!1)
\end{bmatrix}
\!\!=\!\!
\begin{bmatrix}
X(k) \!+\! \tau_s U(k) \\
Y(k) \!+\! \tau_s V(k) \\
Z(k) \!+\! \tau_s W(k) \\
U(k) \\
V(k) \\
W(k)
\end{bmatrix}
\!\!+\!\!
\begin{bmatrix}
0.5 \tau_s^2 F_x(k)/m \\
0.5 \tau_s^2 F_y(k)/m \\
0.5 \tau_s^2 F_z(k)/m \\
\tau_s F_x(k)/m \\
\tau_s F_y(k)/m \\
\tau_s F_z(k)/m
\end{bmatrix}\!\!.
\end{equation}

\HJfinal{The proposed experiment considers a $20$ kg UAV operating inside a moderate-volume UTM corridor composed of four orthogonal rectangular zonotopes, simulating a metropolitan urban canyon restricted to street overflight and vertically bounded between minimum obstacle clearance and a $120$ m ceiling. The initial state and target are $x_0 = [20,\, 0,\, 60,\, 0,\, 0,\, 0]^\top$ and $y_t = [120,\, 20,\, 50]^\top$, respectively. The predictive planner uses a horizon $N = 5$ with a sampling time $\tau_s = 0.5$ s, and cost weights defined as $Q = I_6$, $R = 0.25 I_3$, and $k_s = 50$. Operational constraints restrict velocities to $\pm 20$ m/s in all axes, with horizontal and vertical force limits bounded at $\pm 33$ N and $\pm 66$ N. Finally, the mixed-integer formulation employs a Big-M constant $M = 15000$ and two interpolation points ($N_p = 2$) for the offset-cost path. In this scenario, the midpoint verification is sufficient to capture the orthogonal corridor geometry without unnecessarily increasing the number of binary variables.}

\begin{figure}[htb]
    \centering
    
    \begin{minipage}[b]{0.99\linewidth}
        \centering
        \includegraphics[width=\linewidth]{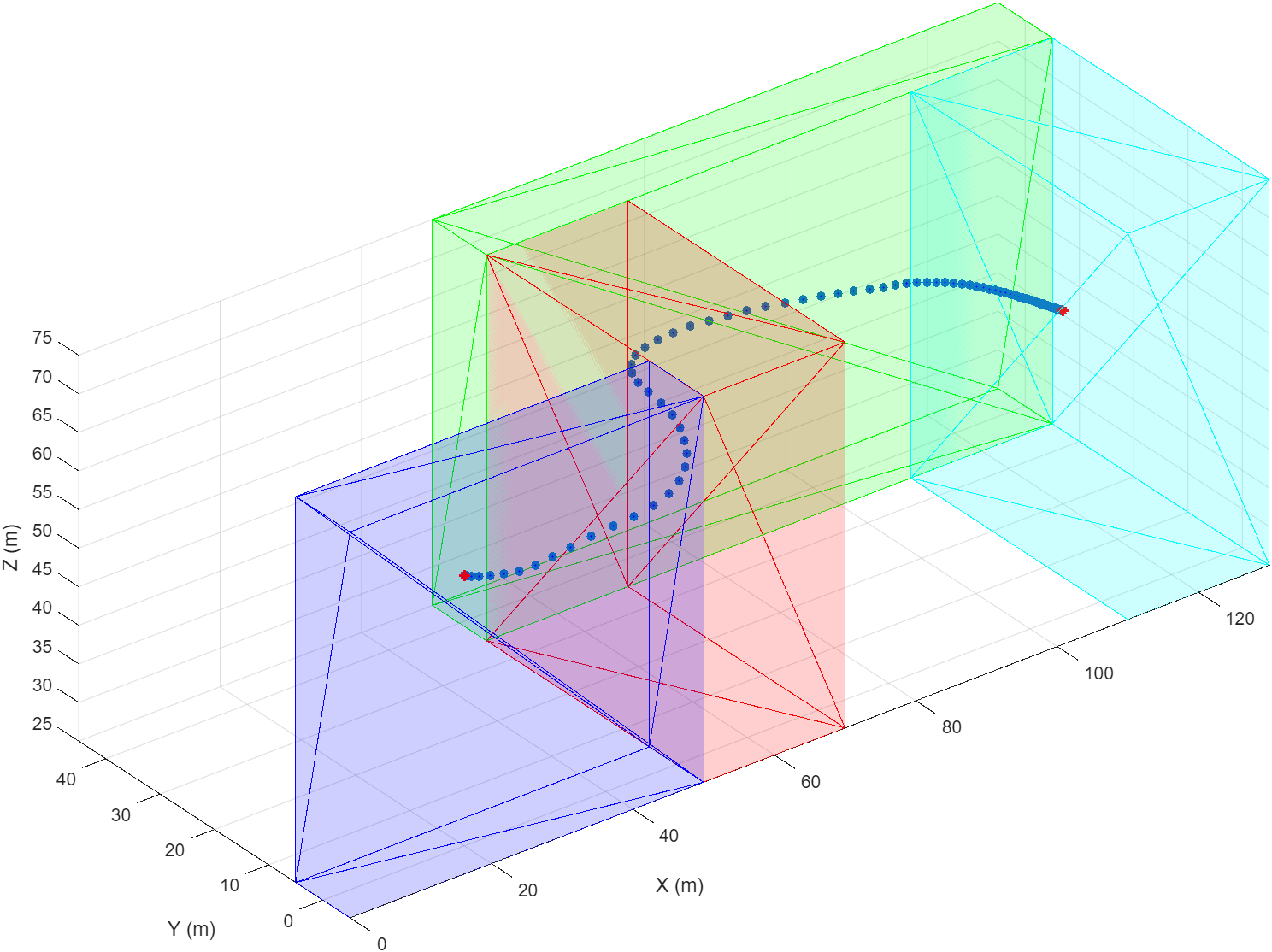}
        \centerline{\footnotesize (a) Isometric View}
    \end{minipage}
    
    \vspace{0.3cm} 
    
    \begin{minipage}[b]{0.99\linewidth}
        \centering
        \includegraphics[width=\linewidth]{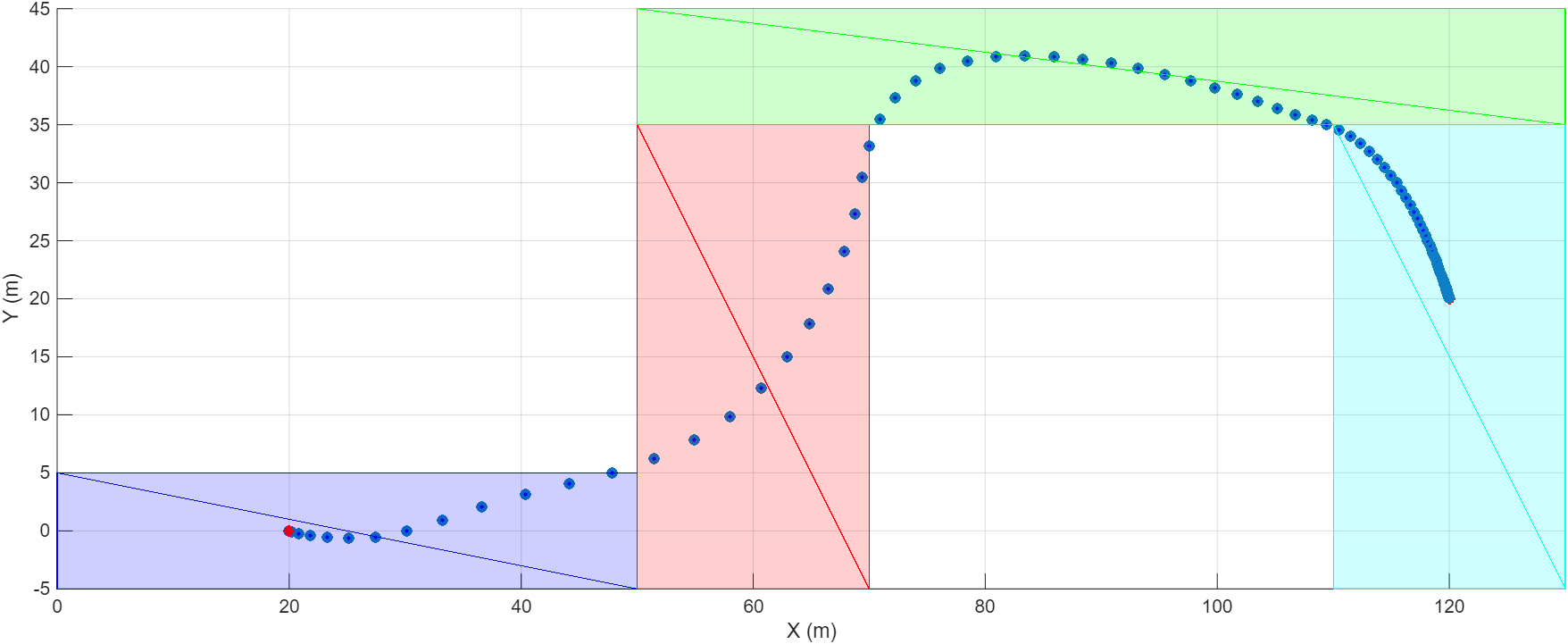}
        \centerline{\footnotesize (b) Top-down View}
    \end{minipage}
    
    \caption{3D and XY projections of the planned trajectory inside the air corridor.}
    \label{fig:trajectory}
\end{figure}

\begin{figure}[!t]
    \centering
    \includegraphics[width=0.95\linewidth]{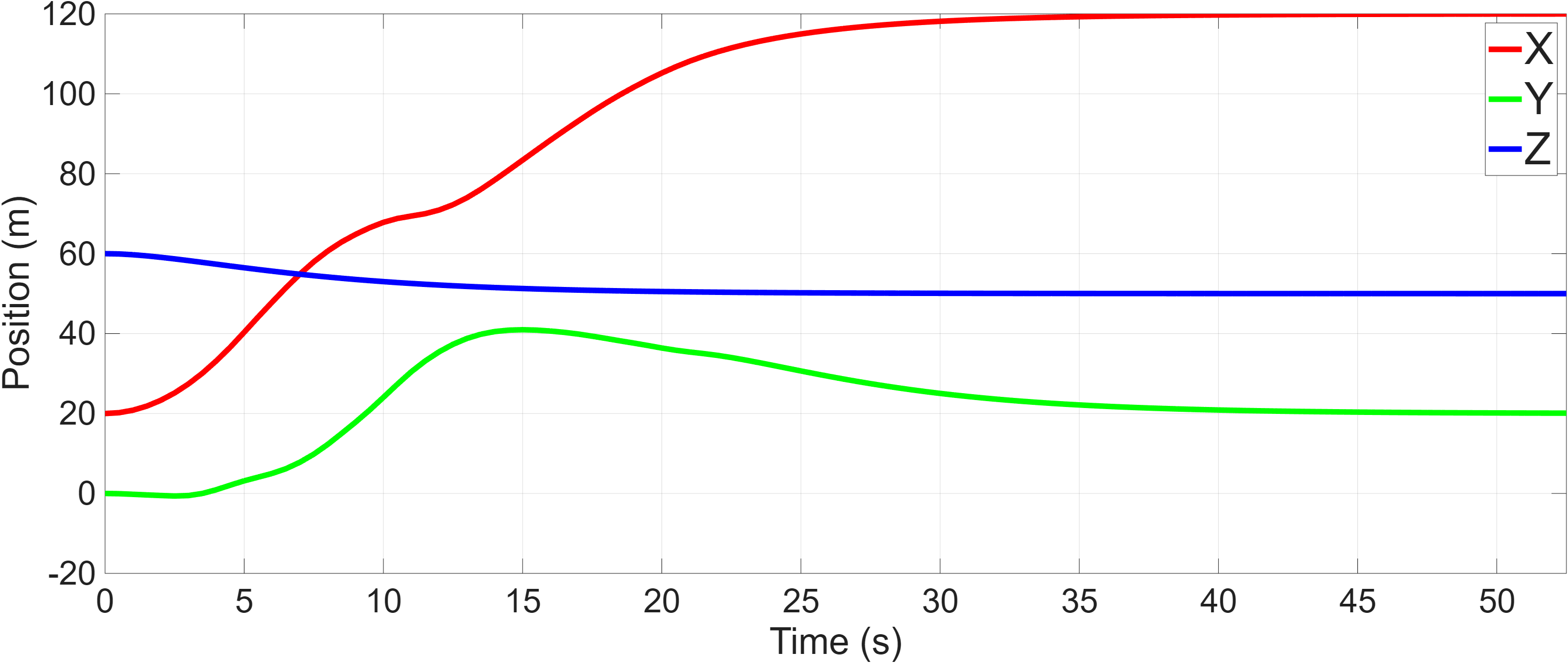}
    \caption{Position evolution of the planned trajectory.}
    \label{fig:position}
\end{figure}

The simulation results demonstrate the efficacy of the proposed formulation in planning a trajectory to guide the UAV through the non-convex corridor. As shown in Figs.~\ref{fig:trajectory} and~\ref{fig:position}, the planner generates a smooth path that converges to the target while avoiding local minima, even with a short prediction horizon ($N = 5$). Furthermore, the mixed-integer formulation manages the switching between corridor segments effectively, ensuring that the predicted states remain within the admissible airspace.

\begin{figure}[htb]
    \centering
    \includegraphics[width=0.95\linewidth]{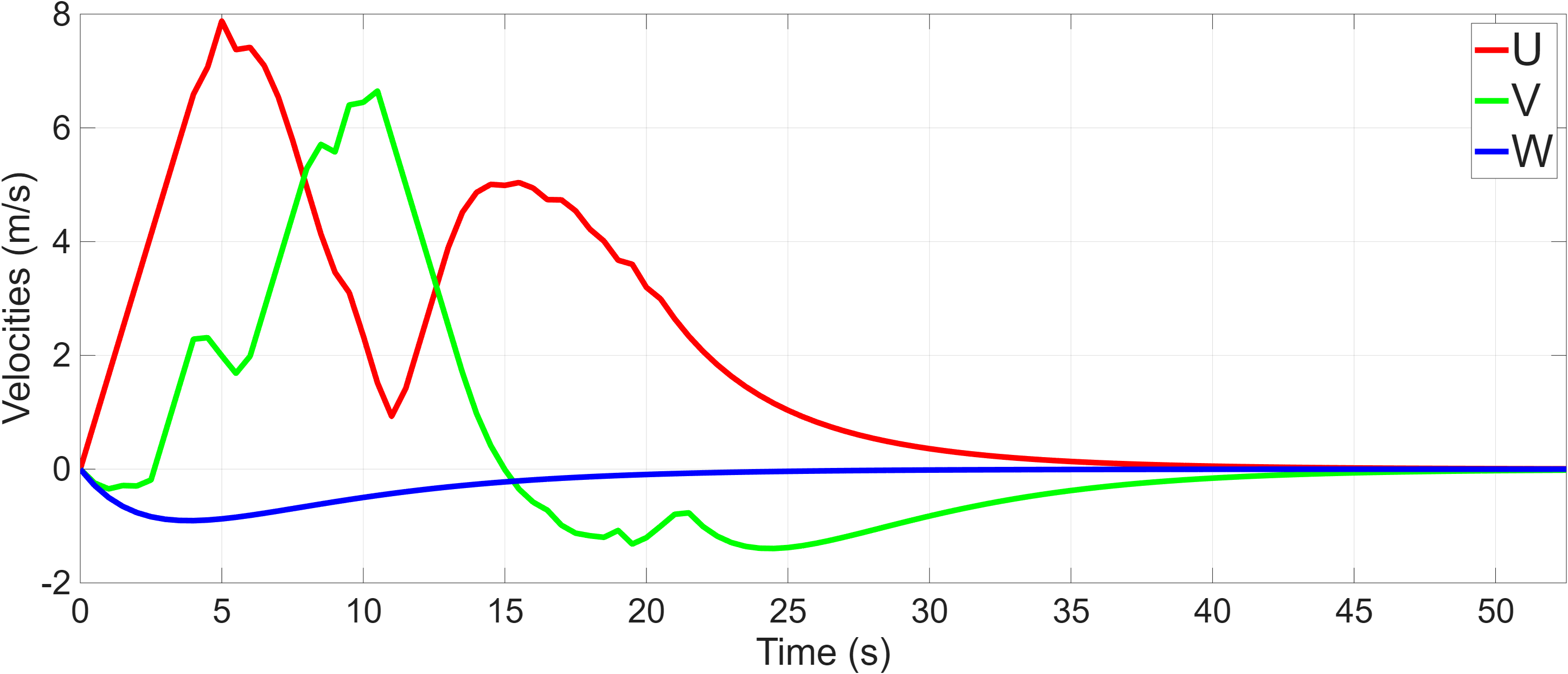}
    \caption{Evolution of the planned velocities.}
    \label{fig:velocities}
\end{figure}

\begin{figure}[htb]
    \centering
    \includegraphics[width=0.95\linewidth]{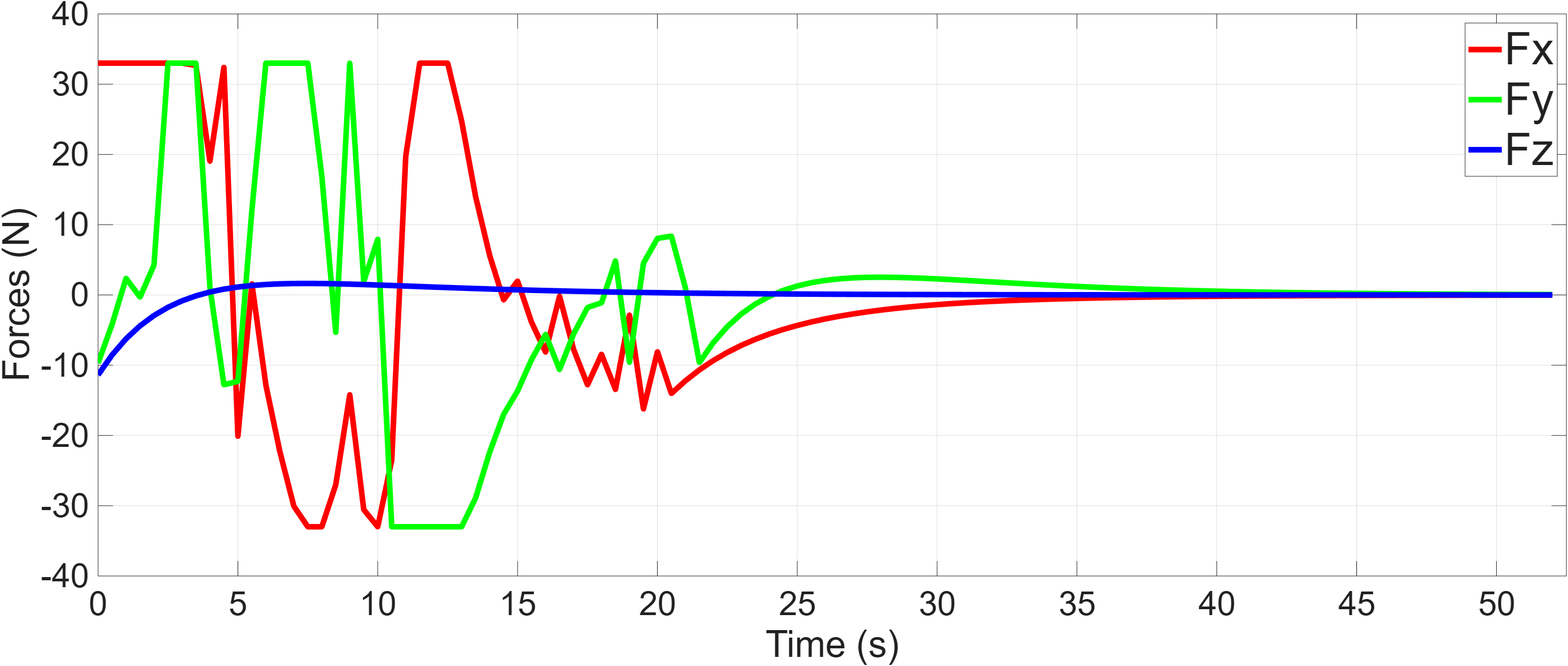}
    \caption{Evolution of the planned forces.}
    \label{fig:forces}
\end{figure}

The velocity profiles shown in Fig.~\ref{fig:velocities} confirm that the UAV operates within the prescribed limits ($\pm 20$ m/s), with smooth transitions during direction changes, particularly in the XY-plane maneuvers. The force inputs in Fig.~\ref{fig:forces} exhibit aggressive behavior and frequently become active at the input constraints ($\pm 33$ N). This behavior follows from the weighting choice in $R$, which penalizes input magnitude but not input variation. If required, the predictive planner can be extended in a straightforward manner to include input-variation constraints, thereby moderating these transitions.


\HJfinal{While the trajectory is dynamically feasible, the relatively large sampling time ($\tau_s = 0.5$ s) adopted to improve prediction capabilities may induce inter-sample constraint violations (tunneling effect) near boundaries, which can be mitigated by introducing an appropriate safety margin in the corridor representation. Furthermore, the mixed-integer formulation's computational burden scales with the chosen design parameters, requiring careful tuning to maintain tractability.}


\section{Conclusion}
\HJfinal{This work presented a motion planning framework for UAV navigation in non-convex urban air corridors. By leveraging a mixed-integer tracking MPC and a shortest-path offset cost, the approach enforces dynamic feasibility and ensures global target convergence without relying on external roadmaps. While simulations demonstrate constraint-compliant trajectories within short horizons, the formulation's integer variables introduce a computational burden tied to topological complexity. Future work will focus on reducing this complexity using Hybrid Zonotopes and geometric methods for continuous segment verification, extending the framework to multi-agent systems and reactive obstacle avoidance, and evaluating its computational performance against other corridor-capable architectures.}

\begin{small}
  \bibliography{references}   

@misc{FAA_UAM_ConOps_v2_2023,
  title        = {{Urban Air Mobility (UAM) Concept of Operations, v2.0}},
  author       = {{FAA}},
  year         = {2023},
  month        = {apr},
  url          = {https://www.faa.gov/sites/faa.gov/files/Urban-Air-Mobility-Concept-of-Operations-2.0.pdf},
  note         = {Accessed on: November 9, 2025}
}

@article{bemporad1999control,
  title={Control of systems integrating logic, dynamics, and constraints},
  author={Bemporad, Alberto and Morari, Manfred},
  journal={Automatica},
  volume={35},
  number={3},
  pages={407--427},
  year={1999},
}

@misc{EASA_EAR_U-space_2024,
  title        = {{Easy Access Rules for U-space (Regulation (EU) 2021/664)}},
  author       = {{EASA}},
  year         = {2024},
  month        = {may},
  url          = {https://www.easa.europa.eu/en/document-library/easy-access-rules/easy-access-rules-u-space-regulation-eu-2021664},
  note         = {Revision from May 2024. Accessed on: November 9, 2025}
}

@misc{ICAO_UTM_Framework_2023,
  title        = {{Unmanned Aircraft Systems Traffic Management (UTM) – A Common Framework}},
  author       = {{ICAO}},
  year         = {2023},
  edition      = {Third},
  url          = {https://www.icao.int/sites/default/files/sp-files/safety/UA/Documents/UTM%20Framework%20Edition%203.pdf},
  note         = {Accessed on: November 9, 2025}
}

@article{uav_survey,
author = {Quan, Lun and Han, Luxin and Zhou, Boyu and Shen, Shaojie and Gao, Fei},
title = {Survey of {UAV} motion planning},
journal = {IET Cyber-Systems and Robotics},
volume = {2},
number = {1},
pages = {14-21},
year = {2020}
}

@Article{drones6050126,
AUTHOR = {Israr, Amber and Ali, Zain Anwar and Alkhammash, Eman H. and Jussila, Jari Juhani},
TITLE = {Optimization Methods Applied to Motion Planning of Unmanned Aerial Vehicles: A Review},
JOURNAL = {Drones},
VOLUME = {6},
YEAR = {2022},
NUMBER = {5},
}

@book{rawlings2009model,
  title={Model Predictive Control: Theory and Design},
  author={Rawlings, James B. and Mayne, David Q.},
  year={2009},
  publisher={Nob Hill Publishing},
}

@article{Brito2019MPCC,
  author={Brito, Bruno and Floor, Boaz and Ferranti, Laura and Alonso-Mora, Javier},
  journal={IEEE Robotics and Automation Letters}, 
  title={Model Predictive Contouring Control for Collision Avoidance in Unstructured Dynamic Environments}, 
  year={2019},
  volume={4},
  number={4},
  pages={4459-4466},
}

@article{layered,
  title={A quantitative framework for layered multirate control: Toward a theory of control architecture},
  author={Matni, Nikolai and Ames, Aaron D and Doyle, John C},
  journal={IEEE Control Systems},
  volume={44},
  number={3},
  year={2024},
}

@article{limon2018,
  title={Nonlinear {MPC} for tracking piece-wise constant reference signals},
  author={Limon, Daniel and Ferramosca, Antonio and Alvarado, Ignacio and Alamo, Teodoro},
  journal={IEEE Transactions on Automatic Control},
  volume={63},
  number={11},
  pages={3735--3750},
  year={2018}
}

@article{santos2024,
  title={Set-point tracking {MPC} with avoidance features},
  author={Santos, Marcelo Alves and Ferramosca, Antonio and Raffo, Guilherme Vianna},
  journal={Automatica},
  volume={159},
  pages={111390},
  year={2024}
}

@inproceedings{tracking_tutorial,
  title={Model predictive control for tracking using artificial references: Fundamentals, recent results and practical implementation},
  author={Krupa, Pablo and K{\"o}hler, Johannes and Ferramosca, Antonio and Alvarado, Ignacio and Zeilinger, Melanie N and Alamo, Teodoro and Limon, Daniel},
  booktitle={Proceedings of the IEEE 63rd CDC},
  pages={2977--2991},
  year={2024},
}

@ARTICLE{homeomorfismo,
  author={Cotorruelo, Andres and Ramirez, Daniel R. and Limon, Daniel and Garone, Emanuele},
  journal={IEEE Transactions on Automatic Control}, 
  title={Nonlinear {MPC} for Tracking for a Class of Nonconvex Admissible Output Sets}, 
  year={2021},
  volume={66},
  number={8},
  pages={3726-3732}}

@ARTICLE{exploration,
  author={Soloperto, Raffaele and Mesbah, Ali and Allgöwer, Frank},
  journal={IEEE Transactions on Automatic Control}, 
  title={Safe Exploration and Escape Local Minima With Model Predictive Control Under Partially Unknown Constraints}, 
  year={2023},
  volume={68},
  number={12},
  pages={7530-7545},
}

@ARTICLE{soloperto,
  author={Soloperto, Raffaele and Köhler, Johannes and Allgöwer, Frank},
  journal={IEEE Transactions on Automatic Control}, 
  title={A Nonlinear {MPC} Scheme for Output Tracking Without Terminal Ingredients}, 
  year={2023},
  volume={68},
  number={4},
  pages={2368-2375},
}

@inproceedings{nascimento2023nmpc,
  author    = {Nascimento, Iuro B. P. and Rego, Brenner S. and Pimenta, Luciano C. A. and Raffo, Guilherme V.},
  booktitle = {Proceedings of the IEEE 62nd CDC},
  title     = {{NMPC} Strategy for Safe Robot Navigation in Unknown Environments using Polynomial Zonotopes},
  year      = {2023},
  pages     = {7100-7105},
}

@article{milagroso,
  title={An {MPC} framework for efficient navigation of mobile robots in cluttered environments},
  author={K{\"o}hler, Johannes and Zhang, Daniel and Soloperto, Raffaele and Carron, Andrea and Zeilinger, Melanie},
  journal={arXiv preprint arXiv:2509.15917},
  year={2025}
}

@article{karur2021survey,
  title={A survey of path planning algorithms for mobile robots},
  author={Karur, Karthik and Sharma, Nitin and Dharmatti, Chinmay and Siegel, Joshua E},
  journal={Vehicles},
  volume={3},
  number={3},
  pages={448--468},
  year={2021},
}

@inproceedings{combastel2013quadratic,
  title={A quadratic programming-based method for determining the distance between a point and a zonotope},
  author={Combastel, Christophe},
  booktitle={Proceedings of the IEEE 52nd CDC},
  pages={655--660},
  year={2013},
}

@inproceedings{althoff2010efficient,
  title={An efficient algorithm for computing the H-representation of a zonotope},
  author={Althoff, Matthias},
  booktitle={Proceedings of the IEEE 49th CDC},
  pages={6739--6744},
  year={2010},
}
\end{small}            
\end{document}